\documentclass[conference]{IEEEtran}
\IEEEoverridecommandlockouts
\usepackage{cite}
\usepackage{amsmath,amssymb,amsfonts}
\usepackage{algorithmic}
\usepackage{graphicx}
\usepackage{textcomp}
\usepackage{booktabs}
\usepackage{xcolor}
\usepackage{subcaption}

\usepackage[table]{xcolor}
\def\BibTeX{{\rm B\kern-.05em{\sc i\kern-.025em b}\kern-.08em
    T\kern-.1667em\lower.7ex\hbox{E}\kern-.125emX}}

\usepackage{tikz}
\usetikzlibrary{shapes.geometric, arrows, positioning}

\begin{document}

\title{Human Activity Recognition Method for Moderate Violence Detection}
\author{
    Luis Angel Aparicio Borjas$^*$, 
    V{\'i}ctor El{\'i}as Nieto$^*$, 
    Juan Irving Vasquez$^\dagger$, \\
    Alfonso Fernandez-Vazquez$^*$, 
    Gerardo Antonio Alvarez Hernandez$^\dagger$ \\
    \small $^*$ESCOM-IPN, Mexico City, M{\'e}xico \\
    \small $^\dagger$Centro de Innovación y Desarrollo Tecnológico en Cómputo, Instituto Politécnico Nacional, Mexico City, M{\'e}xico \\
    \small lapariciob1800@alumno.ipn.mx, veliasn1800@alumno.ipn.mx, jvasquezg@ipn.mx, afernan@ieee.org, galvarezh1400@alumno.ipn.mx
}
\maketitle

\begin{abstract}
Physical violence in public spaces is a significant public health concern, with minor incidents such as pushing often serving as precursors to more severe escalations. This research develops an automated system for the real-time detection of moderate physical violence, specifically pushing, in surveillance camera footage. The proposed solution integrates state-of-the-art computer vision models, utilizing YOLO11 and YOLO11-Pose for human detection and skeletal keypoint extraction. By calculating body inclination and joint angles between shoulders and hips, a Random Forest classifier was trained to distinguish between normal behavior and aggressive physical contact.

The system’s performance was evaluated through three progressive case studies representing increasing levels of difficulty. In controlled environments with frontal views, the model achieved a precision of 0.98. In the most challenging scenario, featuring high-altitude, steep-angle recordings from real-world surveillance infrastructure, the system maintained a precision of 0.72 despite significant perspective distortion and visual noise. 
%Additionally, a graphical user interface (GUI) was implemented to provide real-time alerts and automated video recording of detected incidents. 
These results demonstrate the feasibility of using skeletal analysis for early violence intervention in urban security contexts
\end{abstract}


\section{Introduction}

Violence has been recognized as a global public health crisis since 1996, contributes to increasing homicide rates, and decreases life expectancy. In Mexico, interpersonal violence is particularly prevalent and minor incidents such as pushing often serve as the initial spark for more severe escalations~\cite{Reyes2021NECESIDADES}. Because of human operator fatigue, some acts of moderate aggressions can occur. Traditional surveillance systems struggle to detect those moderate acts of aggression using high volume of visual data \cite{bukht2025review}. Consequently, there is a critical need for automated computer vision solutions that provide early, real-time detection of moderate physical violence to prevent serious conflict and enhance public safety.

Implementing automated detection of moderate physical violence presents a complex challenge in computer vision due to the fine-grained nature of the actions involved. Unlike severe violence or object detection, identifying a "push" requires the precise modeling of human pose dynamics and spatial interactions between multiple subjects. High-altitude surveillance perspectives introduce significant geometric distortions, variable lighting, and frequent occlusions, which complicate the extraction of reliable skeletal keypoints. Furthermore, the system must differentiate between aggressive physical contact and benign urban interactions—such as walking in close proximity or accidental bumping in crowded transit areas—requiring a robust temporal and spatial analysis of joint trajectories and body inclination. Consequently, this problem requires a hierarchical approach that integrates real-time object tracking with precise pose estimation and a classification architecture capable of interpreting subtle biomechanical patterns under diverse environmental constraints.

In light of recent developments in Human Activity Recognition (HAR), significant progress has been made through the adoption of Transformer-based architectures like TransTM \cite{liu2024transtm} and graph convolutional networks such as GLIL \cite{shu2020host}, which excel at capturing long-range dependencies and complex social interactions. These advanced learning methods have shifted the field toward more accurate data-driven and knowledge-driven analysis. However, despite these technical strides and the utilization of datasets like UT-Interaction for identifying aggressive behaviors, most existing models still struggle with low-resolution data, occlusions, and the variability of human movements \cite{bukht2025review}. Consequently, while current methodologies show promise in controlled or specific environments, the problem of robustly detecting pushing and moderate violence in diverse, real-life scenarios remains an unsolved challenge.

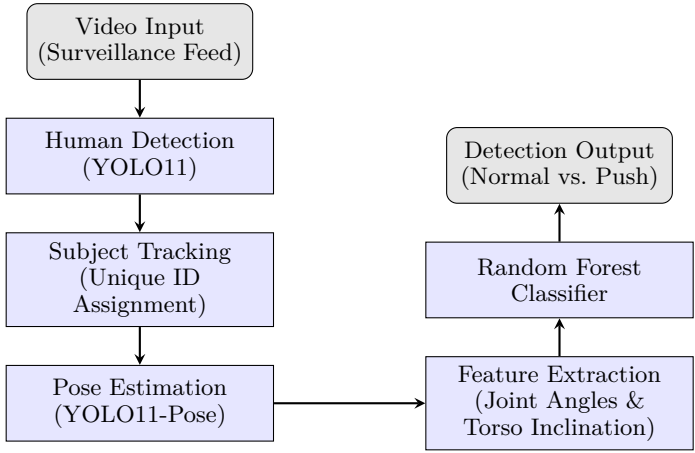
\begin{figure}[tb]
    \centering
    \begin{tikzpicture}[
        node distance=0.5cm,
        every node/.style={font=\small},
        startstop/.style={rectangle, rounded corners, draw, fill=gray!20, text width=2.75cm, align=center, minimum height=1cm},
        process/.style={rectangle, draw, fill=blue!10, text width=3.3cm, align=center, minimum height=1cm},
        decision/.style={diamond, draw, fill=yellow!10, text width=2cm, align=center, inner sep=0pt},
        arrow/.style={thick,->,>=stealth}
    ]

    % Nodes
    \node (input) [startstop] {Video Input \\ (Surveillance Feed)};
    \node (yolo) [process, below=of input] {Human Detection \\ (YOLO11)};
    \node (track) [process, below=of yolo] {Subject Tracking \\ (Unique ID Assignment)};
    \node (pose) [process, below=of track] {Pose Estimation \\ (YOLO11-Pose)};
    \node (features) [process, right=2cm of pose] {Feature Extraction \\ (Joint Angles \& Torso Inclination)};
    \node (rf) [process, above=of features] {Random Forest \\ Classifier};
    \node (output) [startstop, above=of rf] {Detection Output \\ (Normal vs. Push)};

    % Arrows
    \draw [arrow] (input) -- (yolo);
    \draw [arrow] (yolo) -- (track);
    \draw [arrow] (track) -- (pose);
    \draw [arrow] (pose) -- (features);
    \draw [arrow] (features) -- (rf);
    \draw [arrow] (rf) -- (output);

    \end{tikzpicture}
    \caption{General methodology for detection of moderate physical violence using a hierarchical computer vision framework.}
    \label{fig:methodology_flowchart}
\end{figure}

In this study, we propose a method for detecting human-to-human (H2H) violent interactions. See a flow diagram of the method in Fig. \ref{fig:methodology_flowchart}. In particular, we address the challenge of determining whether one person is violently pushing another. Our approach is based on YOLO detection and skeleton extraction; we calculate the relative angles of the subjects and utilize these relative features as inputs for a Random Forest classifier.

The proposed method was tested across three different scenarios, which incrementally increase the level of detection difficulty. The results validate the effectiveness of our approach, demonstrating that it is well-suited for implementation on edge devices.

\section{Related Work}

% Almenos tres párrafos de la revisión de la literatura

Human-to-human (H2H) interaction has been studied for several years now. It remains a challenge as it involves not only detecting the behavior of individuals but also the complex interactions between them \cite{bukht2025review}. Regarding datasets, only a few consider violence, and specifically pushing actions. For instance, \cite{kong2012learning} provides data for H2H interaction where certain violent moves are included, such as boxing and kicking. While \cite{ryoo2010contest} provides examples of pushing, the distance relative to the camera is large, which differs significantly from our target study.

In terms of the techniques applied to H2H interaction, research has transitioned from handcrafted features toward automatic feature extraction and advanced classification techniques, especially deep learning \cite{balakrishnan2024accurate}. Early approaches focused on the use of dynamic texture descriptors and Support Vector Machines (SVMs) to classify actions based on visual appearance \cite{kellokumpu2008human}. As the field evolved, probabilistic methods, such as Hidden Markov Models (HMMs) and Bayesian Networks, were widely adopted to model temporal dependencies and uncertainty in interactive behaviors. By 2017, the integration of 3D skeletal data and depth camera silhouettes became a standard for capturing spatial joint characteristics \cite{manzi2017human}.

Recently, the focus has shifted toward deep learning, specifically the use of Convolutional Neural Networks (CNNs) for spatial feature extraction and Long Short-Term Memory (LSTM) networks to recognize the sequence of actions \cite{ko2018deep}. State-of-the-art models employ transformer-based approaches \cite{liu2024transtm} and graph convolutional networks \cite{shu2020host} to process the extracted features.

\section{Pushing Behavior Detection}

%JI editado
The proposed methodology utilizes a hierarchical framework for real-time human-to-human activity detection, specifically focusing on the detection of one person pushing another (see Fig. \ref{fig:methodology_flowchart}). The process begins with YOLO11 \cite{Ultralytics2025YOLO11} for human detection to obtain bounding boxes for all individuals. Subsequently, YOLO11-Pose is employed to extract 17 skeletal keypoints from the detected subjects \cite{Ultralytics2023Ultralytics}. Unique IDs are assigned to each subject to facilitate consistent tracking across frames. 

%JI editado
The extracted skeleton coordinates are then transformed into biomechanical descriptors, specifically torso inclination and joint angles, calculated through vector dot products to characterize the physical dynamics of a push. Finally, these features are processed by a Random Forest classifier \cite{Saharshla2025Bagging}, which employs an ensemble of decision trees to distinguish between specific behaviors. Details of the methodology are provided below.

% Explicación completa
\subsection{Human detection}

%JI Editado
The human detection is the first stage. The input for this stage consists of frames extracted sequentially from a fixed surveillance camera, which are processed independently.
People are detected with YOLOv11\cite{Ultralytics2025YOLO11}  trained under a single-class configuration to identify only humans under our dataset. This educes the complexity of the model and allows its learning capacity to be concentrated on the visual characteristics of the human body.

For each frame analyzed, the detector generates as output a set of detections consisting of: (i) the detected class (human), (ii) a bounding box that defines the spatial location of the individual within the image, and (iii) a confidence probability associated with each detection. This information is used directly by the subsequent subject tracking module, where the detections serve as the basis for temporal association and the assignment of unique identifiers.

%Innecesario: 
%As an initial component of the proposed system, the human detection module provides reliable spatial localization of individuals in the scene, enabling the correct functioning of the subsequent stages of the pipeline.

\subsection{Subject Tracking}
% JI Editado
The subject tracking module is responsible for maintaining the temporal correspondence of each individual detected throughout the video sequence. This module is activated directly from the output of the human detector, using the tracking mechanism integrated into the YOLOv11.

%Innecesario
%Based on the detections generated in each frame, the tracking module internally assigns a unique identifier (ID) to each detected object, corresponding to a different individual. This identifier is consistently retained over time, allowing each person to be tracked throughout their stay in the scene, even when they move or change position.

%As a result, the system generates a temporal sequence of frames associated with the same individual for each ID, which preserves the subject's identity throughout the video. These sequences constitute a structured representation of each person's movement and are used as input for subsequent stages of the system, in particular for pose estimation and individual behavior analysis.

%In this way, the subject tracking module allows the temporal analysis of human behavior to be decoupled, ensuring that the extracted characteristics and classification decisions are made consistently for the same individual.

\subsection{Pose Estimation}

\begin{figure}
    \centering
    \includegraphics[width=0.95\linewidth]{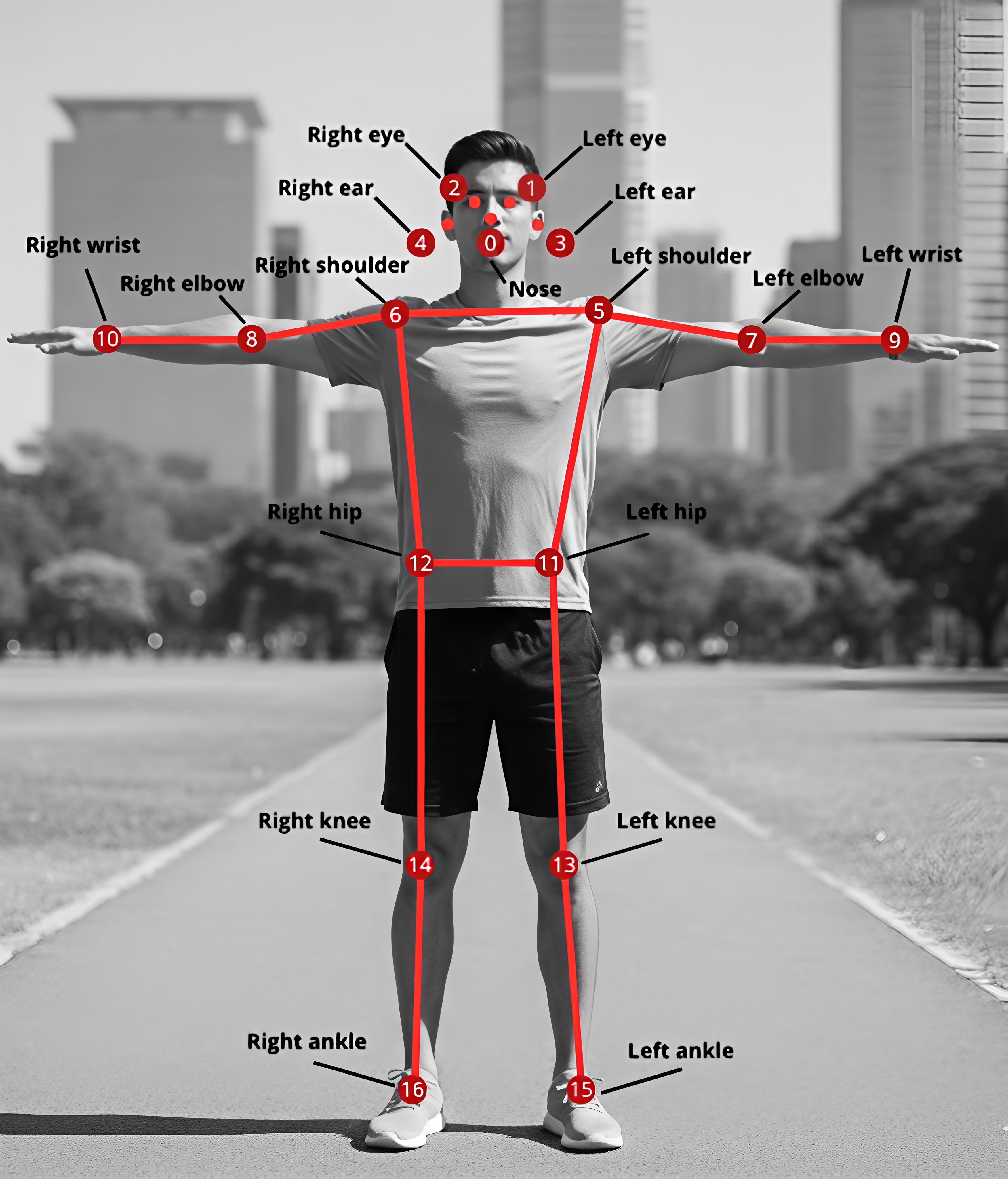}
    \caption{Feature points used for skeleton detection and further activity recognition.}
    \label{fig:keypoints}
\end{figure}

The human pose detection module estimates the body structure of each individual based on the motion sequences generated by the subject tracking module. Pose estimation is performed using the YOLOv11-Pose model, which allows the calculation of keypoints for each frame in the sequence. As output, the model generates a set of 17 keypoints, which represent the individual's body configuration and are shown schematically in Fig. \ref{fig:keypoints}.

%For each unique identifier assigned during tracking, the corresponding frame sequence is delivered as input to this module, allowing for individual and consistent analysis of each person's movement over time~\cite{Ultralytics2023Ultralytics}.

In order to focus the analysis on the body regions most relevant to the study of behavior, only characteristic points numbered from 5 to 17 are used, according to the indexing shown in Fig. \ref{fig:keypoints}. These points correspond mainly to the upper and lower limbs and the torso, discarding those associated with the head and face.  

%The selected characteristic points are stored sequentially and subsequently used in the feature extraction stage, where kinematic and postural patterns associated with each individual's behavior are analyzed.

\subsection{Feature Extraction}

\begin{figure}
    \centering
    \includegraphics[width=0.95\linewidth]{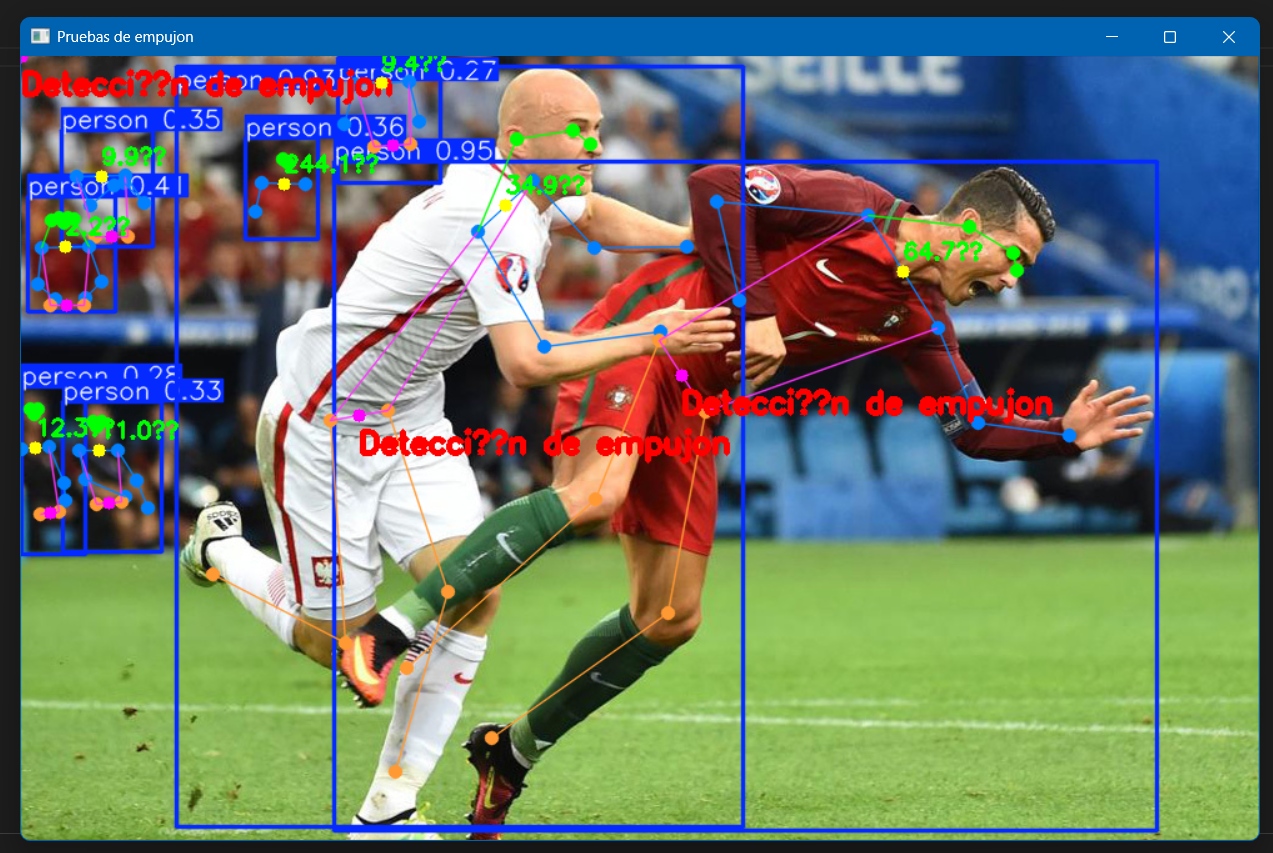}
    \caption{Tests applying the POSE model for identifying pushes}
    \label{fig:empujon}
\end{figure}

Based on empirical observation of human behavior in video surveillance footage, recurring movement patterns associated with pushing events were identified. In particular, it was observed that a pronounced tilt of the shoulder-hip axis may indicate a forced displacement of the body, while an abnormal horizontal extension of the arms is often present during a physical altercation (see Fig. \ref{fig:empujon}). These observations led to the definition of a set of kinematic characteristics aimed at capturing these patterns.

Based on the characteristic points obtained in the pose detection stage, representative vectors of the torso and arms are constructed for each individual and for each frame of the sequence. From these vectors, the angles of inclination of the torso and extension of the arms are calculated using the dot product between vectors, which allows the angular relationship between two body segments to be quantified.

These angles were calculated using the dot product formula for vectors. Given three points A, B, and C (where B is the center point), the vectors $\vec{BA}$ and $\vec{BC}$ are constructed and the following expression is applied:

\begin {equation}
\theta = \cos^{-1} \left( \frac{\vec{BA} \cdot \vec{BC}}{\|\vec{BA}\| \cdot \|\vec{BC}\|} \right).
\end{equation} where:
\begin{itemize}
\item $\vec{BA} \cdot \vec{BC}$ is the dot product between the vectors.
\item $\|\vec{BA}\|$ and $\|\vec{BC}\|$ are the norms (magnitudes) of the vectors.
\item $\theta$ is the resulting angle in radians (later converted to degrees).
\end{itemize}
This formulation allows for a robust measurement of the relative orientation of body segments, regardless of their magnitude. The angles calculated for the torso and arms are encoded as features of interest and stored temporarily for each subject identifier. These features constitute the input representation for the classification module, where the final inference is made to determine whether or not the observed movement corresponds to a pushing~\cite{Gitau2023Using}.

\subsection{Behavior Classification}

The random forest classifier module is responsible for determining whether the behavior observed in the scene corresponds to a pushing event. This module receives as input the kinematic characteristics extracted in the previous stage, which describe the postural and movement dynamics of each individual over time.
A Random Forest model is used for the classification task, which is based on a set of jointly trained decision trees. This approach allows for the internal evaluation of multiple combinations of input characteristics, generating non-linear decision rules that capture complex relationships between movement patterns associated with physical interaction~\cite{Saharshla2025Bagging}.
Since a pushing event necessarily occurs as a result of the interaction between at least two individuals, the classification module incorporates this restriction into the analysis.
 In particular, the classifier's decisions only consider events involving two different people, previously identified by the subject tracking module. This avoids confusion between a push and individual actions that could correspond, for example, to a fall or a spontaneous movement without external interaction.
As an output, the Random Forest module generates a binary decision indicating whether or not the action observed in the video corresponds to a pushing event. This decision is based on the information accumulated throughout the pipeline, including human detection, temporal tracking, pose estimation, and feature extraction, allowing for a formal and consistent interpretation of the observed behavior.

\subsection{Dataset}

\subsubsection{Persons detection}

The dataset used for the person detection task was constructed from frames extracted from video surveillance footage captured by a fixed camera located at a public street (location is omitted on purpose due to security issues). The frames were obtained through temporal sampling of 0.5 seconds, in order to reduce redundancy between consecutive images and ensure adequate variability of the scenes.

A total set of 1,600 images was obtained, which were manually labeled using the LabelImg tool. The annotation process was carried out under a single-class configuration, corresponding to person, with the aim of specializing the training of the YOLOv11 model for the detection of humans in the specific environment of the subway entrance and exit. The labels associated with the annotations are binary, where the value 1 represents the presence of the object of interest (person) and the value 0 corresponds to any other region or object present in the scene. This dataset allowed the detector to be adapted to general real conditions.

\subsubsection{Pose estimation}

\begin{figure}[tb]
    \centering
    \begin{subfigure}{\linewidth}
        \centering
        \includegraphics[width=0.95\linewidth]{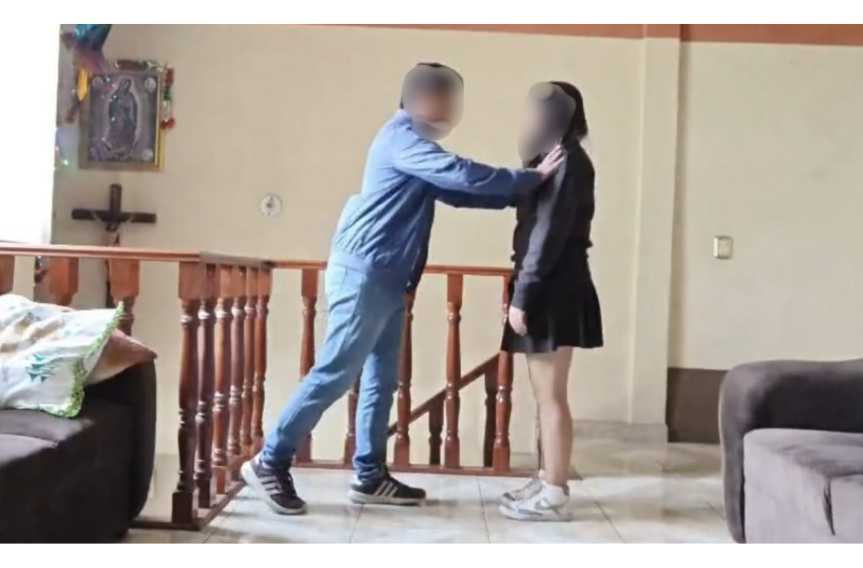}
        \caption{Case 1: Short-distance view with high level of detail}
        \label{fig:case1}
    \end{subfigure}
    
    \begin{subfigure}{\linewidth}
        \centering
        \includegraphics[width=0.95\linewidth]{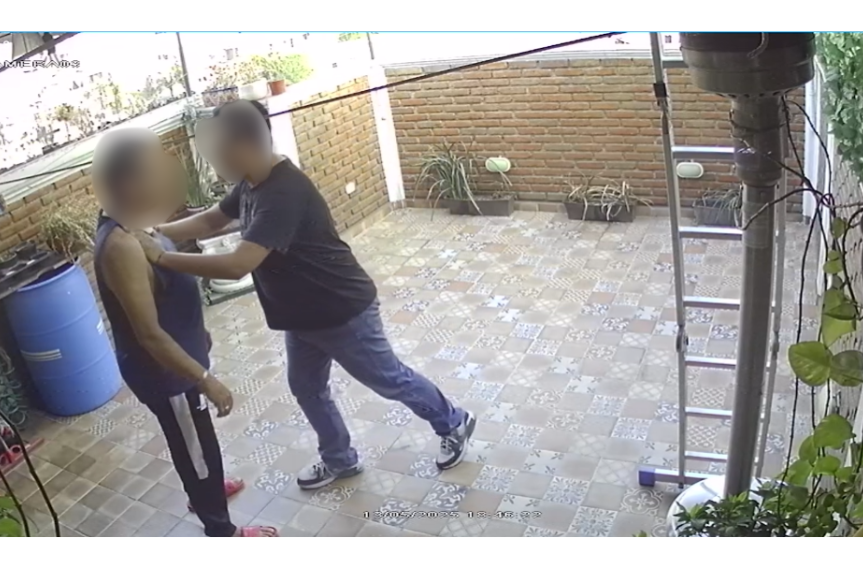}
        \caption{Case 2: Medium-distance view from a considerable height}
        \label{fig:case2}
    \end{subfigure}
    
    \begin{subfigure}{\linewidth}
        \centering
        \includegraphics[width=0.95\linewidth]{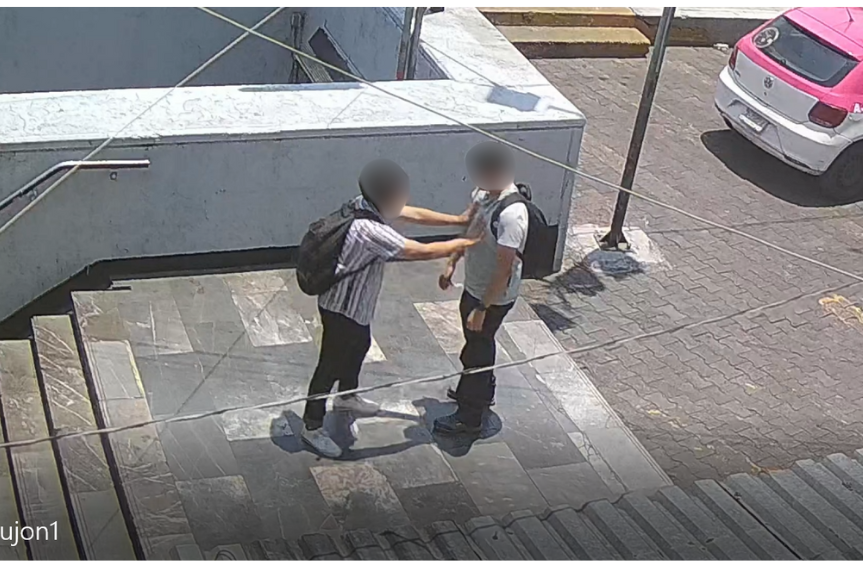}
        \caption{Case 3: Long-distance panoramic view in a realistic surveillance scenario}
        \label{fig:case3}
    \end{subfigure}

    \caption{Examples of the three testing scenarios for detecting human pushing. Faces are hidden on purpose due to privacy concerns.}
    \label{fig:three_images_column}
\end{figure}

In addition to the previous dataset that is used only for training YOLO. The dataset used for the task of classifying actions consists of short video clips showing different types of human movement. See some examples of the test subset in \ref{fig:three_images_column}. Each clip was labeled according to the predominant action, considering only two categories: pushing and normal movement. 

To analyze the module's behavior under different view and scale configurations, the dataset was organized into three independent sets, corresponding to three case studies. The first case considers a camera close to people, where individuals are observed at a short distance and with a high level of detail; the videos in this set have a resolution of $848 \times 848$ pixels and are composed of 45 clips labeled as pushing and 45 clips labeled as normal movement. The second case corresponds to a camera located at a considerable height, where people are observed at medium distances, using a resolution of $1920 \times 1080$ pixels; this set consists of 21 pushing clips and 21 normal movement clips. Finally, the third case represents a realistic video surveillance scenario, with a camera installed at a high altitude and a panoramic view, where people appear at a great distance; this set, also with a resolution of $1920 \times 1080$ pixels, contains 44 clips corresponding to pushing and 44 clips corresponding to normal movement (see Fig \ref{fig:three_images_column}).

% Asegúrate de incluir esto en tu preámbulo:

% \usepackage{graphicx}

From each video clip, the subject tracking and pose detection steps described above were applied, obtaining the selected characteristic points of the human body. Based on these points, four relevant angles defined between the shoulders and hips were calculated, which allow the capture of postures and movements characteristic of a push event. 

%These angles were stored using the following notation:
%$angle\_11\_5\_6, angle\_12\_6\_5, angle\_11\_6\_12$, and $angle\_12\_5\_11$.
%The extracted features were stored in CSV files, where each record corresponds to a video clip. In total, 28 features were used, of which 24 correspond to the two-dimensional coordinates (x, y) of characteristic points 5 to 16, following the notation
%$kp5\_x, kp5\_y, …, kp16\_x, kp16\_y$,
%shown in Fig. X, and 4 correspond to the angles calculated from the relative behavior between shoulders and hips. Finally, each record was assigned a label indicating the corresponding class (normal or push), thus forming the dataset used for training and evaluating the Random Forest classifier.

\begin{table*}[tb]
\centering
\caption{Experimental setup and performance results for the evaluated YOLOv11 models.}
\label{tab:yolo11_experiments}
\begin{tabular}{cccccccc}
\toprule
\textbf{ID} & \textbf{Modelo} & \textbf{LR} & \textbf{Batch} & \textbf{Epoch} & \textbf{Precision} & \textbf{Recall} & \textbf{mAP50 / mAP50--95} \\
\midrule
1  & YOLO11s & $1\times10^{-3}$ & 16 & 194 & 0.92883 & 0.88184 & 0.91661 / 0.72037 \\
2  & YOLO11s & $1\times10^{-3}$ & 32 & 250 & 0.93417 & 0.88626 & 0.91442 / 0.71479 \\
3  & YOLO11s & $1\times10^{-4}$ & 16 & 194 & 0.92883 & 0.88184 & 0.91661 / 0.72037 \\
4  & YOLO11s & $1\times10^{-4}$ & 32 & 250 & 0.93417 & 0.88626 & 0.91442 / 0.71479 \\

5  & YOLO11m & $1\times10^{-3}$ & 8  & 162 & 0.90512 & 0.88841 & 0.90590 / 0.71964 \\
6  & YOLO11m & $1\times10^{-3}$ & 16 & 239 & 0.90000 & 0.90556 & 0.91063 / 0.73557 \\
7  & YOLO11m & $1\times10^{-4}$ & 8  & 162 & 0.90512 & 0.88841 & 0.90590 / 0.71964 \\
8  & YOLO11m & $1\times10^{-4}$ & 16 & 239 & 0.90000 & 0.90556 & 0.91063 / 0.73557 \\

%\rowcolor{gray!20}
\textbf{9}  & \textbf{YOLO11l} & $\mathbf{1\times10^{-3}}$ & \textbf{8}  & \textbf{250} & \textbf{0.92629} & \textbf{0.89641} & \textbf{0.92072 / 0.74014} \\

10 & YOLO11l & $1\times10^{-3}$ & 16 & 220 & 0.91513 & 0.90353 & 0.91846 / 0.73171 \\
11 & YOLO11l & $1\times10^{-4}$ & 8  & 250 & 0.92629 & 0.89641 & 0.92072 / 0.74014 \\
12 & YOLO11l & $1\times10^{-4}$ & 16 & 220 & 0.91513 & 0.90353 & 0.91846 / 0.73171 \\
\bottomrule
\end{tabular}
\end{table*}

\section{Experiments}

% Párrafo que describa los experimentos de manera general
The experiments carried out evaluate the performance of the proposed system for detecting pushing events in video surveillance scenarios.

% repite 
%, which is implemented as a hierarchical and sequential pipeline. The system is organized into two main components: the first corresponds to the detection and tracking of people, integrating the human detection and subject tracking modules to identify the individuals present in the scene and maintain their identity over time; the second component is aimed at detecting pushes and includes human pose detection, kinematic feature extraction, and Random Forest classification modules, which are responsible for analyzing movement sequences and determining the occurrence of physical interaction. 

The datasets used were divided according to the classic 80–20 scheme, where 80\% of the data was used to train the models and the remaining 20\% was used to validate and test the trained model (10\% respectively).

% Párrafo de implemencación
The implementation and evaluation were carried out on a workstation equipped with an Intel Core i7-10700 processor at 2.90 GHz, 32 GB of RAM at 2933 MHz, and an NVIDIA GeForce RTX 2080 SUPER graphics card with 8 GB of dedicated memory. The entire pipeline was developed using the Python programming language, relying on the PyTorch framework for the implementation and execution of deep learning models, as well as for the integration of the different system modules.

\subsection{Person detection}

With the aim of determining the most suitable architecture for the YOLOv11 model for the task of detecting people in video surveillance scenarios, different configurations were evaluated, considering the trade-off between model complexity and performance. In particular, three variants of the architecture were analyzed: YOLOv11-Small, YOLOv11-Medium, and YOLOv11-Large, which differ in the number of parameters and their representation capacity.

For each architecture, a systematic search of hyperparameters was performed using a grid search strategy. The hyperparameters evaluated included two learning rate values ($1\times 10^{-3}$ and $1\times10^{-4}$), a maximum of 250 training epochs, and an early stopping criterion when no improvement in performance was observed for 50 consecutive epochs. Regarding the batch size, values of 8 and 16 were used for the Medium and Large versions, while batch sizes of 16 and 32 were used for the Small version, considering its lower computational demand.

As a result, a total of 12 experiments were carried out. The performance of each configuration was evaluated using standard object detection metrics: precision, recall, mAP@50, and mAP@50–90, the results of which are summarized in Table \ref{tab:yolo11_experiments}. According to these metrics, the best performance was obtained in the experiment with ID 9, which was selected for integration into the complete pipeline.

% repetitivo
%Once the person detection model was trained, the subject tracking module integrated into YOLOv11 was activated. This module is used as a preprocessing stage, allowing the generation of coherent temporal sequences for each detected individual, which constitute the input for the subsequent stage of pose detection and action classification.

\begin{figure}
    \centering
    \includegraphics[width=1.0\linewidth]{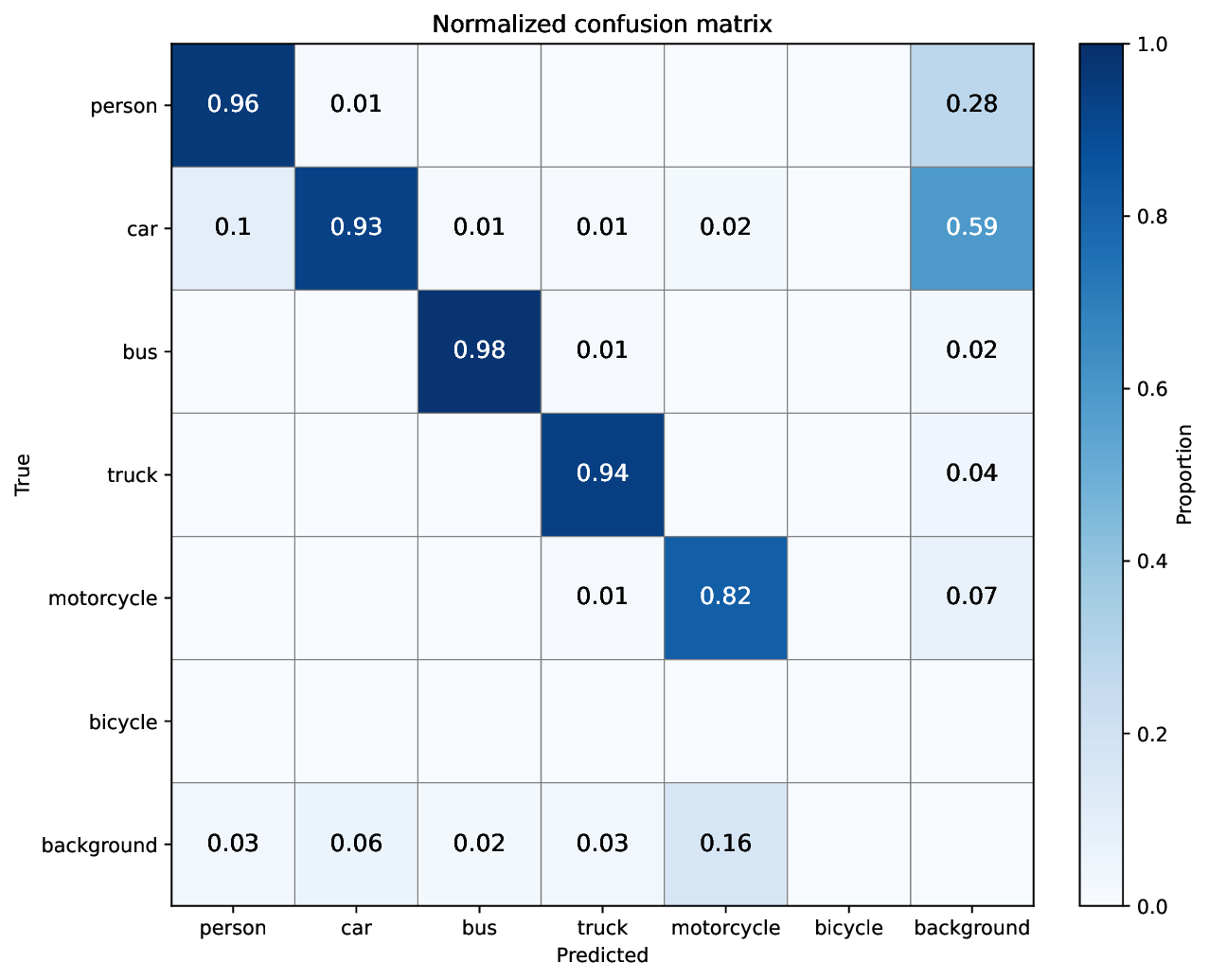}
    \caption{Normalized confusion matrix for experiment ID 9.}
    \label{fig:matrizyolo}
\end{figure}

%Additionally, in order 
To evaluate the detector's generalization ability in environments not seen during training, a confusion matrix (see Fig. \ref{fig:matrizyolo}) was calculated considering, in addition to the person class, other classes that the YOLO model can inherently infer, such as car, bus, truck, and motorcycle. This analysis yielded an overall accuracy of 0.96, indicating that the model was correctly adapted to the new video surveillance scenario and performed adequately for use within the proposed system.

\subsection{Pose Estimation}

For the final stage of the proposed pipeline, corresponding to pose detection and action classification, the YOLOv11-Pose model was used exclusively as a feature extractor, so it was not retrained. 

% Redundante
%Instead, the pre-trained weights provided by the model were used directly, taking advantage of its ability to robustly estimate characteristic points of the human body in video sequences.

%From the video clips processed in the subject tracking stage, YOLOv11-Pose generates human poses frame by frame, which form the basis for the feature extraction process described above. 

%At this stage, the estimated poses are not used for direct classification, but rather function as an intermediate representation that allows the calculation of the kinematic features necessary for behavior analysis.

%These features are then used as input for the Random Forest-based classification model. 

The Random Forest classifier was trained using the following hyperparameters: 100 decision trees, a reproducibility seed equal to 42, a minimum number of samples to split a node of 2, and a minimum number of samples per leaf of 1. 
 
%This configuration allows nonlinear relationships between the extracted features to be captured, while maintaining a balance between generalization capacity and model complexity.

\section{Analysis}

In order to evaluate the performance of the system under different levels of input data complexity, three independent experiments were conducted, each corresponding to one of the previously defined case studies. In the tables \ref{tab:conf_case1}, \ref{tab:conf_case2}, \ref{tab:conf_case3}, shows the confusion matrices obtained for the final classification in each experiment, which allow us to analyze the behavior of the system and its ability to distinguish between normal movements and push events in different video surveillance scenarios.

\begin{table}[tb]
\centering
\caption{Normalized Confusion Matrix – Case 1}
\label{tab:conf_case1}
\renewcommand{\arraystretch}{1.2}
\begin{tabular}{c cc}
\toprule
\textbf{True / Pred} & Normal & Push \\
\midrule
Normal & \textbf{0.91} & 0.09 \\
Push   & 0.08 & \textbf{0.92} \\
\bottomrule
\end{tabular}
\end{table}

\begin{table}[tb]
\centering
\caption{Normalized Confusion Matrix – Case 2}
\label{tab:conf_case2}
\renewcommand{\arraystretch}{1.2}
\begin{tabular}{c cc}
\toprule
\textbf{True / Pred} & Normal & Push \\
\midrule
Normal & \textbf{1.00} & 0.00 \\
Push   & 0.25 & \textbf{0.75} \\
\bottomrule
\end{tabular}
\end{table}

\begin{table}[tb]
\centering
\caption{Normalized Confusion Matrix – Case 3}
\label{tab:conf_case3}
\renewcommand{\arraystretch}{1.2}
\begin{tabular}{c cc}
\toprule
\textbf{True / Pred} & Normal & Push \\
\midrule
Normal & \textbf{0.69} & 0.31 \\
Push   & 0.22 & \textbf{0.78} \\
\bottomrule
\end{tabular}
\end{table}

The experimental results show that the proposed system performs adecuately in both stages of the pipeline. In human detection, the evaluated YOLOv11 models achieved high accuracy and mAP values (0.92), confirming their ability to adapt to the video surveillance environment under consideration, even in the presence of non-target classes. 
%This performance made it possible to generate reliable detections and coherent temporal sequences, which constitute an adequate basis for subsequent behavior analysis. 
In the pose detection and action classification stage, the results obtained show correct discrimination between normal movements and pushing events in the first two case studies, where lighting, scale, and perspective conditions favor accurate estimation of human pose. However, in case 3, corresponding to a scenario of greater visual complexity, a decrease in classification performance is observed, mainly attributable to factors such as lighting variations, greater distances between the camera and the subjects, and more pronounced perspectives. These conditions affect the quality of the pose estimation and, consequently, the discriminability of the extracted kinematic features. 

Despite these limitations, the results confirm the viability of the proposed approach and highlight the importance of considering environmental conditions in the automatic analysis of human behavior.

\section{Conclusions}

This research addressed the challenge of detecting moderate physical violence in video surveillance, specifically focusing on the detection of violent pushing between individuals. We proposed and validated a hierarchical computer vision framework that integrates real-time object detection and pose estimation via YOLO11 with a Random Forest classifier trained on extracted relative kinematic features.

The experimental results confirmed that the proposed pipeline performs effectively under various constraints. The human detection stage achieved a high mean Average Precision, providing a reliable foundation for subject tracking and pose analysis. While the system demonstrated high classification accuracy in scenarios with favor lighting and scale (Case 1 and Case 2), the evaluation of Case 3 revealed that performance is sensitive to extreme camera distances and adverse lighting. These findings highlight that while posture-based features like joint angles and torso inclination are robust indicators of aggressive intent, environmental complexity remains a significant bottleneck for pose-based HAR systems.

Therefore, our method proves to be a viable and adequate solution for implementation on edge devices, offering a scalable alternative to human-monitored surveillance. 

Future work will explore the integration of temporal dependencies through architectures such as LSTMs or Transformers to enhance the system's robustness against the environmental variables encountered in unconstrained urban settings. As well as adding more human activities.

\section*{Acknowledgment}
Authors thank to SECIHTI, IPN-SIP Research Projects 20250144 and 20250342. 

\bibliographystyle{unsrt}
\bibliography{bibliografia} 

\end{document}